\documentclass{article}

\usepackage[preprint]{neurips_2023}

\usepackage[utf8]{inputenc} 
\usepackage[T1]{fontenc}    
\usepackage{hyperref}       
\usepackage{url}            
\usepackage{booktabs}       
\usepackage{amsfonts}       
\usepackage{nicefrac}       
\usepackage{microtype}      
\usepackage{xcolor}         
\usepackage{bbm}
\usepackage{bm}
\usepackage{float}
\usepackage{booktabs}
\usepackage{amssymb}
\usepackage{mathtools}
\usepackage{bbding}
\usepackage{algorithm}
\usepackage{algpseudocode}
\usepackage{subcaption,caption, comment}
\algblock{Input}{EndInput}
\algnotext{EndInput}
\algblock{Output}{EndOutput}
\algnotext{EndOutput}

\newtheorem{theorem}{Theorem}[section]

\numberwithin{equation}{section} \numberwithin{theorem}{section}

\def\P{ {\mathbb P}}                  \def\E{ {\mathbb E}}
                   
\def\R{ {\mathcal R}}              \def\X{ {\mathcal X}}

\def\N{ {\mathcal N}}

\def\D{ {\mathcal D}}

\def\Pr{\operatorname{Pr}}
\def\H{\mathcal{H}}

\title{Conformal Inference for Invariant Representation}

\author{%
Wenlu TANG \\
Department of Applied Mathemetics\\
Hong Kong Polytechnic University\\
\texttt{wenlu.tang@polyu.edu.hk} \\
\And
Zicheng LIU\\
Department of Statistics\\
The Chinese University of Hong Kong\\
\texttt{1155061903@link.cuhk.edu.hk} \\
}

\begin{document}

	\maketitle

	\begin{abstract}	
		The application of machine learning models can be significantly impeded by the occurrence of distributional shifts, as the assumption of homogeneity between the population of training and testing samples in machine learning and statistics may not be feasible in practical situations.
		One way to tackle this problem is to use invariant learning, such as invariant risk minimization (IRM), to acquire an invariant representation that aids in generalization with distributional shifts. 
		This paper develops methods for obtaining distribution-free prediction regions to describe uncertainty estimates for invariant representations, accounting for the distribution shifts of data from different environments.
		Our approach involves a weighted conformity score that adapts to the specific environment in which the test sample is situated. We construct an adaptive conformal interval using the weighted conformity score and prove its conditional average under certain conditions.	
		To demonstrate the effectiveness of our approach, we conduct several numerical experiments, including simulation studies and a practical example using real-world data.
	\end{abstract}
	\section{Introduction}
	The assumption of independent and identically distributed training and test data is a fundamental principle of statistical machine learning. However, in real-world scenarios, distributional shifts are common and can pose challenges for machine learning models when applied to new data (\cite{sugiyama2007covariate}, \cite{taori2020measuring}). To address these challenges, researchers have proposed various strategies to develop predictive models that can adapt to changes in the distribution of the data. The negative impact of such shifts on the quality of predictions can be significant, and it is therefore essential to construct prediction intervals that not only provide point prediction but also quantify the uncertainty of predictions under distributional shifts.
	
	Conformal inference has become a powerful and valuable tool for constructing prediction intervals. Conformal prediction is a distribution-free approach that constructs prediction intervals with a guaranteed coverage probability in finite samples. The objective is to create a prediction set $C_{n}$ that includes $Y_{n+1}$ with a probability of at least $1-\alpha$ (\cite{vovk2005algorithmic}). Since the pioneering work by \cite{vovk2005algorithmic}, there have been numerous follow-up studies and extensions in both computation and theory. \cite{lei2013distribution}, \cite{lei2014distribution}, and \cite{lei2018distribution} have developed much insightful work on statistical theory for conformal methods. Marginal validity, a conventional coverage guarantee that can be achieved under the i.i.d assumption, is demonstrated in \cite{vovk2005algorithmic} and \cite{zeni2020conformal}. However, as demonstrated in \cite{lei2014distribution} and \cite{vovk2012conditional}, conditional validity with a finite-length prediction interval is impossible without regularity and consistency assumptions on the model and estimator. Recently, \cite{romano2019conformalized} and \cite{kivaranovic2020adaptive} proposed a technique that combines conformal prediction with classical quantile regression. \cite{lei2018distribution} prove that conformal prediction intervals are accurate under minimal assumptions on the residuals, indicating that they do not substantially over-cover. Besides the theoretical results, the computational efficiency of the conformal method is also attractive. The mentioned conformal methods rely on the assumption of sample exchangeability, but distribution shifts are prevalent, which can violate this assumption.
	
	
	Recently, numerous studies in conformal inference have aimed to extend the method to better adapt to distribution shifts.  \cite{gibbs2021adaptive}constructed prediction sets in an online setting where the data generating distribution can change over time in an unknown way. Meanwhile, \cite{zaffran2022adaptive} analyzed the impact of this method in time series. \cite{tibshirani2019conformal} proposed a method to obtain accurate prediction sets for covariate-shifted data by weighting calibration data using the likelihood ratio, while  \cite{barber2022conformal} used a non-symmetric algorithm that assigns more weight to recent observations. \cite{cauchois2020robust} examined distribution shift in an $f$-divergence ball and proposed using distributional robust optimization to create prediction sets.  \cite{guan2023localized} expanded the conformal prediction framework by offering a locally-focused, single-test-sample adaptive construction that can be adapted to various conformal scores. \cite{dunn2022distribution} extended conformal methods to a hierarchical setting by using data from multiple training domains. 
	
	This paper aims to quantify the uncertainty estimates of distribution shift through out-of-distribution generalization via Invariant Risk Minimization. This approach assumes that the variation across the training domain represents the variation in the testing domain.
	
	\subsection{Invariant Risk Minimization}
	Even though neural networks can perform exceptionally well on the training data by Empirical Risk Minimization (ERM), where the average loss is considered across all training environments, they can be highly sensitive to distributional shifts, which makes it challenging to apply them in practical situations. Therefore, it is essential to focus on identifying and addressing distributional shifts to improve the performance of machine learning models in real-world applications. Invariant risk minimization (IRM) (\cite{arjovsky2019invariant}), is  one of the approaches aims to learn models that are robust to distributional shifts in the data.
	
	Let us suppose data are coming from a collection of $m$ environments denoted by $\mathcal{E}_{tr}$, which includes environments $e\in\{1, \ldots, m\}$, each defined on the sample space $\mathcal{X} \times \mathcal{Y}$. Here, $\mathcal{X}$ and $\mathcal{Y}$ denote the input and target spaces, respectively. The joint distribution  in distinct environments is denoted by $\Pr^e(\boldsymbol{x}, \boldsymbol{y})$, where $e \in \mathcal{E}$. Let $\mathcal{D}^e:=\left\{\left(\mathbf{x}_i^e, \mathbf{y}_i^e\right)\right\}_{i=1}^{n_e}$ be the data set drawn from $e \in \mathcal{E}_{t r}$ with $n_e$ being data set size. Suppose that data $\D^e=(x_i^{e},y_i^{e})_{i=1}^{n_e}$ is collected from different environments $ e\in\{1, \ldots, m\},$ and $n_e$ is the sample size in each environment.
	The basic idea behind IRM is to identify the underlying causal mechanisms without environment spurious effect and to explicitly model them. This allows the model to generalize across different environments, even if they have different distributions. Essentially, the goal is to learn a model that is invariant to changes in the data distribution. The main invariance assumption in IRM is the existence of a data representation $\Phi(X)$ such that for all $e, e^{\prime} \in \mathcal{E}_{t r}$, $\mathbb{E}\left[Y \mid \Phi\left(X^e\right)\right]=\mathbb{E}\left[Y \mid \Phi\left(X^{e^{\prime}}\right)\right]$, where $\mathcal{E}_{ {tr }}$ denotes the available training environments. To achieve this goal, IRM uses a regularized optimization approach that encourages the model to learn feature representation $\Phi(x)$ and optimal classifier $\omega$ that are predictive of the target variable, but are also invariant to changes in the data distribution. Existing IRM methods learn $\omega$ and $\Phi(x)$ by solving the following optimization problem:
	\begin{align*}
		\min _{\omega, \Phi} \sum_{e\in \mathcal{E}_{tr}} \mathcal{R}^e(\omega, \Phi)+\lambda\mathcal{J}(\omega, \Phi),
	\end{align*}
	where $\mathcal{R}^e(\omega, \Phi)=\E_{X^e,Y^e}[l(\omega \cdot \Phi(X^e),Y^e)]$ is the risk under environment $e$ and $l$ is the loss function. The regularizer penalty $\mathcal{J}(\omega, \Phi)$ is designed to ensure that the model learns features that are common across all environments and it varies in different approaches. In \cite{arjovsky2019invariant}, they propose  {\it IRMv1} where $\mathcal{J}(\omega, \Phi)=\left\|\nabla_{\omega} R^e(\omega \cdot \Phi)\right\|^2$. Several studies have suggested variations to this objective by introducing alternative regularizer  to fit stronger invariance assumptions. Under a stronger invariance assumption that $\mathbb{P}[Y \mid \Phi(X^e)]=\mathbb{P}[Y \mid \Phi(X^{e^{\prime}})]$, \cite{krueger2021out} proposed {\it REx} with $\mathcal{J}(\omega, \Phi)=\operatorname{Var}\left(\mathcal{R}^e(\omega, \Phi)\right)$ and \cite{chang2020invariant} proposed {\it InvRat} with  $\mathcal{J}(\omega, \Phi)=\lambda \left(R_e\left(\omega, \Phi\right)-R_e\left(\omega_e, \Phi\right)\right)$. Several follow-up works has been proposed to make the IRM more effective. \cite{zhou2022sparse} add a sparsity constraint to the network and  train a neural network that is sparse to prevent overfitting. \cite{lin2022bayesian} updated {\it InvRat} by Bayesian method with a posterior distribution of the classifier. \cite{rosenfeld2020risks} points out some limitations with IRM on classification tasks. \cite{mahajan2021domain} introduces a novel regularizer to match the representation of the same object in different environments. \cite{wang2022provable} proposes a simple post-processing method for solving the IRM problem without retraining the model. \cite{ahuja2020invariant} searches for Nash equilibrium solutions between several environments in a game and let all the environments play the game to determine the action to minimize overall risk. \cite{chang2020invariant}, \cite{koyama2020out},\cite{li2022invariant} and \cite{ahuja2021invariance} consider the invariant learning problem from the information theory aspect. 
	\cite{creager2021environment} proposed Environment Inference for Invariant Learning(EIIL) that attempts to automatically partition a dataset into different environments to learn environments labels that maximize the IRM’s penalty. IRM has been applied to a variety of machine learning tasks, including classification, regression, and reinforcement learning. It has shown promising results in settings where there are known or unknown distributional shifts in the data, such as in healthcare, finance, and other domains where data is constantly changing.
	
	The methods of IRM vary in terms of objective functions, penalties, training frameworks, and optimization techniques. Although IRM's effectiveness is often evaluated by comparing its classification accuracy on out-of-distribution testing data to that of ERM, there is currently no specific criterion to assess the performance of the invariant representation. Furthermore, there is limited literature on the development of valid prediction intervals for invariant learning problems. These gaps in the literature motivate the present study.

	\subsection{Summary and Outline}
	In this paper, we make the following methodological and theoretical contributions.
	\begin{itemize}
		\item We propose a criterion to evaluate the invariance performance of the invariant representation obtained in Invariant Risk Minimization (IRM). We demonstrate that the expectation of an ideal invariant predictor across different environments only varies due to covariate shift.
		\item We perform split conformal inference on training data from various environments by combining the data from each environment to construct a conformal interval. We then evaluate the marginal coverage for this interval.
		\item We propose a weighted conformity score, where the weight of a data point $x_i$ corresponding to environment $e$ is determined by the similarity of $x_i$ and samples in environment $e$. We then construct an adaptive conformal interval using this weighted conformity score, which holds locally coverage in each environment.
		\item We establish theoretical guarantees for the valid coverage of our proposed approach and provide a method to check the condition using the invariance assessment criterion.  Extensive numerical studies are conducted to support the theory.
		
		{\small\begin{table}[H]
				\caption{A comparison of some recent conformal interval under distribution shift.}
				\label{evaluate}
				\centering
				\begin{tabular}{cc|ccc}
					\toprule
					Literature&Method     & Multi-environment&Online  & Conditional validity   \\
					\midrule
					\cite{dunn2022distribution}	 & Two-Layer  &\CheckmarkBold &   \XSolidBrush  &\XSolidBrush\\
					\cite{gibbs2021adaptive}     & ACI & \XSolidBrush&   \CheckmarkBold   &\XSolidBrush\\
					\cite{cauchois2020robust}    & DRO &\XSolidBrush&\XSolidBrush& \XSolidBrush\\
					This paper & ACIR&\CheckmarkBold &\CheckmarkBold & \CheckmarkBold    \\
					\bottomrule
				\end{tabular}
		\end{table}}
	\end{itemize}
	In Section 2.1, we present a novel statistic for the invariant representation that can assess the invariance of the representation. Section 2.2 introduces the split conformal method for data from different environments, while Section 2.3 proposes a weighted conformity score based on the similarity between the testing point and each environment. This weighted conformity score is adaptive to the variation in different environments, and we use it to construct an adaptive conformal prediction interval for the invariant predictions, which we provide theoretical guarantees for. In Section 3, we conduct extensive simulation studies to examine the finite-sample performance of the proposed method. Finally, we provide concluding remarks in Section 4, and all technical proofs are deferred to the Supplementary Materials.

	\section{Methodology}
	In this section,  we propose a split conformal and adaptive conformal interval for IRM predictions.
	\paragraph{Preliminaries} Suppose that training data $\D^e=(x_i^{e},y_i^{e})_{i=1}^{n_e}, $ is collected from different environments $e=1, \ldots, m$ with joint distribution $\Pr^e(x,y)$ in the environment $e$, where $n_e$ is the sample size in each data set $\D^e$.  Then {\it IRMv1} (\cite{arjovsky2019invariant}) solves 
	\begin{equation}\label{IRMv1}
		\min_{\Phi, \omega} \sum_{e \in \mathcal{E}_{\mathrm{tr}}} \R^e(\Phi,\omega)+\lambda\left\|\nabla_{w \mid w=1.0} R^e(w , \Phi)\right\|^2,
	\end{equation}
	where $\R^e(\Phi,\omega)$ is the risk w.r.t the environment $e$,$\Phi: \X \to \H$ is the data representation mapping covariates from covariate domain $\X$ to the embedded space $\H$, and $\omega: \H\to \mathcal{Y}$ is the classifier in classification or the last layer in multilayer perception in regression problems. We can obtain the  estimate for the invariant predictor $\E(Y|X=x)$, which is  $\hat{f}(x)=\hat{\omega} \cdot \hat{\Phi} (x)$.
	
	\subsection{Invariance Assessment}
	In \cite{arjovsky2019invariant}, the key idea of the invariance assumption is that the expected value of the outcome variable given the invariant representation $\Phi(\cdot)$ is identical across environments. Specifically, it assumes that there exists invariant representation $\Phi(\cdot)$, such that for distinct $e, e'\in \mathcal{E}_{tr}$, $\mathbb{E}\left[Y \mid \Phi\left(X^e\right)\right]=\mathbb{E}[Y \mid \Phi(X^{e^{\prime}})]$, where $X^e$ is the covariate collected from environment $e$.  If we take the expectation w.r.t. $X^e$ on both sides, we get:
	\begin{align*}
		&\E_{X^e}\left\{\E\left[Y \mid \Phi\left(X^e\right)\right]\right\}=\int_{\X}\E\left[Y \mid \Phi\left(X^e\right)\right]d\P(X^e)\\
		=&\int_{\X}\mathbb{E}[Y \mid \Phi(X^{e^{\prime}})]d\P(X^e)
		=\int_{\X}\mathbb{E}[Y \mid \Phi(X^{e^{\prime}})]d\P(X^{{e^\prime}})\frac{d\P(X^{{e}})}{d\P(X^{{e^\prime}})}\\
		=&\E_{X^{e^{\prime}}}\left\{\mathbb{E}[Y \mid \Phi(X^{e^{\prime}})]\rho(X^{{e^\prime}},X^{e})\right\},
	\end{align*}
	where $\rho(X^{e'}, X^{e})$ is the the likelihood ratio of $d\P(X^{{e}})/d\P(X^{{e^\prime}})$. 
	The invariance assumption posits that the expected value of the outcome variable given the invariant representation $\Phi(\cdot)$ is the same across different environments. This implies that the transformation of the mean value of the invariant representation $w \cdot \Phi(\cdot)$ across environments only depends on the covariate shift $\rho(X^{e'}, X^e)$. Previous studies have investigated the covariate shift $\rho(X^{{e'}}, X^{e})$, see \cite{sugiyama2005input}, \cite{quinonero2008dataset}, \cite{chen2016robust}. If we know or can accurately estimate the likelihood ratio between every pair of environments, we can evaluate the performance of the invariant representation $\Phi(\cdot)$. For instance, we can choose a baseline environment $e$ and compute 
	$$M_{e}(\Phi, X^{e'}) = \E_{X^{e^{\prime}}}\left[\mathbb{E}[Y \mid {\Phi}(X^{e^{\prime}})]\rho(X^{{e^\prime}},X^{e})\right]$$ 
	for $e' = 1, \ldots, m$ and $\rho(X^{{e}},X^{e})=1$. It can be estimated by empirical expectation 
	$$\hat{M}_{e}(\Phi, X^{e'})=\sum_{x_i: i\in e'}\hat{\omega}\cdot \hat{\Phi}(x_i)\rho(X^{{e'}}, X^{e}).$$ 
	The variance of $M_{e}(\hat{\Phi}, X^{e'})$ can serve as a benchmark for the invariance of the estimated invariant representation $\hat{\Phi}(\cdot)$. Denote 
	\begin{equation}\label{inv}
		{\text{Inv}}(\Phi)=\E_e\{Var_{e'}(M_{e}(\Phi^{or}, X^{e'}))\}
	\end{equation}
	as the average of the variance across the environment. For the optimal invariant representation $\Phi^{or}(\cdot)$ that satisfies invariance assumption, ${\text{Inv}}(\Phi^{or})=0$. If the estimate from the IRM model approximates the optimal invariant representation $\Phi^{or}$, the value ${\text{Inv}}(\Phi)$ will be small. Table \ref{evaluate1} shows the value of ${\text{Inv}}(\Phi)$ in the SEM setting same as the setting in \cite{arjovsky2019invariant}. It shows that the ${\text{Inv}}(\Phi)$ value in IRM model is smaller than the ERM model.
	\begin{table}[H]
		\caption{The invariance evaluation by ${\text{Inv}}(\hat{\Phi})$ value.}
		\label{evaluate1}
		\centering
		\begin{tabular}{cllll}
			\toprule
			Method     & FOU     & POU &FEU &PEU\\
			\midrule
			ERM &  0.067 & 0.094&0.131& 0.184    \\
			IRM  & 0.045 &0.052 &0.122&  0.161   \\
			\bottomrule
		\end{tabular}
	\end{table}

	\subsection{Split Conformal Prediction Interval}
	In this section, we want to quantify the uncertainty of the estimates and conduct a distribution-free prediction interval for given miscoverage rate $\alpha$ using the invariant prediction and conformal inference. By solving (\ref{IRMv1}), we obtain the invariant estimates $\hat{f}(x)=\hat{\omega} \cdot \hat{\Phi} (x)$ for any $x\in \X$. 
	
	We begin by applying the split conformal method to the data from different environments, similar to method 1 in \cite{dunn2022distribution}. Specifically, we split the data in each training set into $\D_{tr}^e$ and $\D_{cal}^e$, where $\D_{cal}=\{\D_{cal}^e\}_{e=1}^m$ and $\D_{tr}=\{\D_{tr}^e\}_{e=1}^m$, with sample sizes $n_{cal}^e$ and $n_{tr}^e$, respectively. We train the model (\ref{IRMv1}) and obtain $\hat{f}(x)$ using data in $\D^e_{tr}$ for $e=1,\ldots,m$. The conformity score function is defined by $S(x,y)=|\hat{f}(x)-y|$, and we use it to compute the conformity score $\varepsilon_{i}^e=S(y_i^e, x_i^e)$ for $i\in \D_{cal}$ and $e=1, \ldots, m$. We then rank all the $\varepsilon_{i}^e=S(y_i^e, x_i^e)$ in ascending order and define the $100(1-\alpha)\%$ empirical quantile of $\varepsilon_{i}^e$ as $Q_{(1-\alpha)}(\varepsilon_{i}^e, \D_{cal})$. Finally, we define a prediction interval as 
	\begin{equation}\label{SC}
		\hat{C}_{1-\alpha}^{SC}(x)=\{y: S(y, x)\leq Q_{(1-\alpha)}(\varepsilon_{i}^e, \D_{cal})\}.
	\end{equation}
	In each environment, we assume that $S(X^e, Y^e)\sim \P_S^e$. Since the joint distribution of $\P_{XY}^e$ is not identical across environments, the distribution $\P_S^e$ also differs. As shown in \cite{dunn2022distribution}, when we pool data from different environments and denote the joint distribution by $\P_{XY}^1,\ldots, \P_{XY}^m\sim \Pi$, a new observation $(x_{n+1},y_{n+1})$ collected from a new distribution $P_{XY}^{m+1}\sim \Pi$ satisfies $(x_{n+1},y_{n+1})\sim \widetilde{\Pi}$, where $\widetilde{\Pi}=\int P d \Pi(P)$. The constructed $\hat{C}^{SC}(x)$ defined in (\ref{SC}) satisfies the marginal coverage for new observations.
	\begin{theorem}
		For any new observation $(x_{n+1},y_{n+1})$ from $\widetilde{\Pi}$, it holds that $$\P\left(y_{n+1}\in \hat{C}_{1-\alpha}^{SC}(x)\right)\geq 1-\alpha.$$
	\end{theorem}
	The above result indicates that while $(x_i^{e},y_i^{e})$ are not exchangeable across environments, they are independently sampled and not identical. To ensure marginal coverage on the pooled data distributions, we compute the average of the pooled joint distributions $(X,Y)$ and the empirical quantile on the pooled residuals across environments. 
	
	\paragraph{Remark} In addition to the proposed approach, \cite{dunn2022distribution} propose alternatives, such as subsampling approaches, to construct prediction sets for observations from distinct distributions.  These methods aim to recover the population distribution by either pooling all subpopulations or subsampling across the training groups. However, a limitation of the subsampling method is that marginal coverage holds only when the number of environments $m > 1/\alpha - 1$, which requires large group sizes to achieve a small miscoverage rate. To address this limitation, we propose an adaptive conformal method that does not require the number of groups and is adaptive to shifts among the groups.

	\subsection{Calibrated Prediction Inverval}
	Now we construct a conformal interval adapt to the distribution shifts across the environments.
 We first fit the IRM model (\ref{IRMv1}) using the training data $\D_{tr}$ to obtain estimates of $\hat{\omega}$ and $\hat{\Phi}$. Next, we use the conformity score function $S(x,y)=|\hat{f}(x)-y|$ to compute the conformity score $\varepsilon_{i}^e=S(y_i^e, x_i^e)$ for $i\in \D_{cal}$ and $e=1, \ldots, m$. In contrast to split conformal method in Section 2.2 where all data was pooled together, we now weight the conformity score for each environment $e$ separately, following the approach in \cite{guan2023localized}. To construct a weighted conformity score, we assign weights to new observations $(x_{n+1},y_{n+1})$ based on the data in the calibration set $\D_{cal}$. Specifically, we compute the weight for any new input $x_{n+1}$ by considering the first and second moment differences of the samples, similar to the idea of moment matching in \cite{arbel2017moment}. We calculate the mean and standard deviation of each invariant representation sample $\hat{\Phi}(x_i)$ in the calibration set ${\D_{tr}^e}_{e=1}^m$, denoted by $\mu(\hat{\Phi}(x_i))$ and $V(\hat{\Phi}(x_i))$, respectively, i.e.,
	\begin{equation}\label{ave}
		\mu(\hat{\Phi}(x_i))=d^{-1}\sum_{j=1}^{d}\hat{\Phi}(x_{ij}), V({\hat{\Phi}(x_i)})= \{\hat{\Phi}(x_{ij})-\mu(\hat{\Phi}(x_i))\}^{1/2},
	\end{equation} 
	for $i\in \{\D_{tr}^e\}_{e=1}^m.$ Then we calculate the mean and variance of each invariant representation sample $\hat{\Phi}(x_i)$ in the calibration set ${\D_{tr}^e}_{e=1}^m$ to obtain $\mu^e(\hat{\Phi})=(n^e_{cal})^{-1}\sum_{i=1}^{n^e_{cal}}\mu({\hat{\Phi}(x_i)})$ and $V^e(\hat{\Phi})=(n^e_{cal})^{-1}\sum_{i=1}^{n^e_{cal}}V({\hat{\Phi}(x_i)})$, which represent the average mean and variation in environment $e$. For a new input $x_{n+1}$, we define $\tau(x_{n+1}, e)$ as the similarity between $x_{n+1}$ and the samples in environment $e$ based on their sample mean and variance differences. Specifically, $\tau(x_{n+1}, e)$ is computed as 
	\begin{equation}\label{tau}
		\tau(x_{n+1}, e)=\exp(-|V(\hat{\Phi}(x_{n+1}))-V^e(\hat{\Phi})|)\exp(-|\mu(\hat{\Phi}(x_{n+1}))-\mu^e(\hat{\Phi})|),
	\end{equation}
    We then assign weights $w_{x_{n+1},e}$ to sample $x_{n+1}$ with respect to environment $e$, which are constructed using the similarity measure $\tau(x_{n+1}, e)$ and normalized by the sum of similarities across all environments, i.e.,
    \begin{equation}\label{weight}
    	w_{x_{n+1},e}=\frac{\tau(x_{n+1}, e)}{\sum_{e=1}^{m}\tau(x_{n+1}, e)}.
    \end{equation}
	 Let $\hat{Q}_{(1-\alpha)}^e(\varepsilon_{i}^e, \D_{cal}^e)$ be the $100(1-\alpha)\%$ empirical quantile of $\{\varepsilon_{i}^e\}_{i=1}^{n^e_{cal}}$ in the $e$ environment. 
	 Next, we compute the weighted conformity score using the weights $w_{x_{n+1}, e}$ and the $100(1-\alpha)\%$ empirical quantile $\hat{Q}{(1-\alpha)}^e(\varepsilon{i}^e, \D_{cal}^e)$ of ${\varepsilon_{i}^e}{i=1}^{n^e{cal}}$ in environment $e$. Specifically, the weighted conformity score is defined by
	 \begin{equation}\label{conformity}
	 	\tilde{Q}_{(1-\alpha)}(x_{n+1})=\sum_{e=1}^{m}\hat{Q}_{(1-\alpha)}^e(\varepsilon_{i}^e, \D_{cal}^e)\times w(x_{n+1}, e).
	 \end{equation}
   This $\tilde{Q}_{(1-\alpha)}(x_{n+1})$ weights conformity score in each environment according to the the sample $x_{n+1}$. Finally, the adaptive conformal interval for invariant representation (ACIR) is constructed as 
   \begin{equation}\label{ACIR}
   	\widehat{C}^{ACIR}_{1-\alpha}=\left[\hat{\omega}\cdot\hat{\Phi}(x_{n+1})-\tilde{Q}_{(1-\alpha)}(x_{n+1}), \hat{\omega}\cdot\hat{\Phi}(x_{n+1})+\tilde{Q}_{(1-\alpha)}(x_{n+1})\right],
   \end{equation}
    where $\tilde{Q}{(1-\alpha)}(x_{n+1})$ represents the data-driven weight function of $x_{n+1}$ that measures the discrepancy between the true value of $y_{n+1}$ and invariant prediction $\hat{\omega}\cdot\hat{\Phi}(x_{n+1})$. Algorithm 1 shows the implementation of the ACIR construction.
	
	

	\begin{algorithm}\label{ACIR_algo}
		\caption{The Adaptive Conformal Interval for Invariant Representation(ACIR)}
	\hspace*{\algorithmicindent} \textbf{Input:}  Data $(x_i^e, y_i^e)$ from $e=1,\ldots, m$ environments, $x_{n+1}$, the IRM model $\mathcal{A}$, and $\alpha\in (0,1)$\\
	\hspace*{\algorithmicindent} \textbf{Output:} A prediction interval $\widehat{C}(x_{n+1})$ for unobserved $y_{n+1}$. \\ 
	\begin{algorithmic}[1]
		\State Randomly split the training data into 2 subsets $\D_{tr}^e, \D_{cal}^e$ with sample size $n_{tr}^e, n_{cal}^e$ in each environment.
		\State Train $\mathcal{A}$ on all samples in $\{\D_{tr}^e\}_{e=1}^m$: $\hat{\omega}, \hat{\Phi}\leftarrow \mathcal{A}\left(\{(x_i,y_i)\}_{i\in \{\D_{tr}^e\}_{e=1}^m}\right)$.
		\State Compute $\varepsilon_{i}^e=|y_i^e-\hat{\omega}\hat{\Phi}(x_i^e)|$ for each $i\in \D_{tr}^e$ and $e\in \{1,\ldots, m\}$.
		\State  Compute $\hat{Q}_{(1-\alpha)}^e(\varepsilon_{i}^e, \D_{cal}^e)$ as the $\lceil (1-\alpha(1+n_{cal}^e)) \rceil$-th largest value in  $\{\varepsilon_{i}^e\}_{i=1}^{n^e_{cal}}$ in the $e$ environment. 
		\State Calculate the mean and standard deviation of each sample point in every environment and determine the environment-wise average $\{V^e(\hat{\Phi})\}_{e=1}^m$ and $\{\mu^e(\hat{\Phi})\}_{e=1}^m$ using equation (\ref{ave}).
		
		\State  For a new input $x_{n+1}$, compute the weight $	w_{x_{n+1},e}$ according to equation (\ref{weight}).
	\State  Compute the weighted conformity score $\tilde{Q}_{(1-\alpha)}(x_{n+1})$ applying equation (\ref{conformity}).
		\State Use $\tilde{Q}_{(1-\alpha)}(x_{n+1})$ to construct the prediction set at $x_{n+1}$ as:
		$\widehat{C}^{ACIR}_{1-\alpha}=\left[\hat{\omega}\cdot\hat{\Phi}(x_{n+1})-\tilde{Q}_{(1-\alpha)}(x_{n+1}), \hat{\omega}\cdot\hat{\Phi}(x_{n+1})+\tilde{Q}_{(1-\alpha)}(x_{n+1})\right]$.
	\end{algorithmic}
	\end{algorithm}
 The proposed  prediction interval $\widehat{C}^{ACIR}_{1-\alpha}(x)$ is adaptive to changes in data distribution over time when operating with online streaming data. When new data $x_{t+1}$ is received, we assign a weighted conformity score to the observation and construct a conformal interval that adapts to shifts among the environments.
\begin{theorem}
	If $\|\hat{\Phi}-{\Phi}^{or}\|_{\infty}=o_p(1)$, where ${\Phi}^{or}$ is the oracle $\Phi$ that satisfies invariance assumption, and new observation $(x_{n+1}^e,y_{n+1}^e)$ is from $\widetilde{\Pi}$, it holds that $$\P\left(y^e_{n+1}\in \widehat{C}_{1-\alpha}^{ACIR}(x^e_{n+1})\mid x^e_{n+1}\right)= 1-\alpha+o_p(1).$$
\end{theorem}
Our results demonstrate that, under standard regularity conditions, the conformal prediction set is nearly optimal and may provide asymptotic conditional coverage, particularly when the initial model estimator is accurate. This finding is consistent with previous studies  \cite{lei2014distribution} and \cite{lei2018distribution}.
However, in practice, the oracle representation $\Phi$ is unknown. Therefore, we use the statistic ${\text{Inv}}(\Phi)$ defined by (\ref{inv}) to check the invariance assumption. Furthermore, we can calibrate the invariant prediction by calculating the average distinction of the invariance in the environment $e$ using $\delta^e=|(m-1)^{-1}\sum_{i\neq e}\hat{M}_{e}(\Phi, X^{i})-\hat{M}_{e}(\Phi, X^{e})|$. When $\Phi=\Phi^{or}$, $\E(\delta^e)=0$, and $\E(\delta^e)$ increases as $\|\Phi-\Phi^{or}\|_{\infty}$ increases. Therefore, we can add $\sum_{e=1}^mw(x_{n+1},e)\delta^e$ when constructing the conformal interval (\ref{ACIR}) to calibrate $\hat{\omega}\cdot\hat{\Phi}$ according to the invariance. In our numerical experiments, we observed that the values of $\delta^e$ were consistently small in relation to weighted conformity scores. As a result, we decided to omit this step from our analysis presented in Section 3.


	
	\section{Numerical results}
	In this section, we apply the proposed ACIR interval to the synthetic data and a open-source financial data. We show the advantages of the proposed ACIR compare to the split conformal without weight conformity score in both ERM and IRM estimates.
	\subsection{Synthetic data}
	Here we present experiments on the (linear) structural equation model (SEM) tasks introduced by \cite{arjovsky2019invariant} and \cite{krueger2021out}. We first simulate a synthetic dataset with a $10$-dimensional vector of $X$ that contains causal effects $X_1$ and non-causal effect $X_2$,  and a continuous response $Y$ from the distributions as below.
	\begin{align*}
		H^e&\sim \N(0, e^2)\\
		X_1 & \sim \N(0, e^2)+W_{H \rightarrow 1} H^e \\
		Y & \sim W_{1 \rightarrow Y} X_1+ \N(0, \sigma_y^2)+ W_{H \rightarrow 1} H^e \\
		X_2 & \sim W_{Y \rightarrow 2} Y+\N(0,\sigma_2^2)
	\end{align*}
	The variance of these distributions may vary across domains.  We consider the following two cases, 
	\begin{enumerate}
		\item Fully-observed (F), where $W_{h \rightarrow 1}=W_{h \rightarrow y}=W_{h \rightarrow 2}=0$, or partially-observed (P), where $\left(W_{h \rightarrow 1}, W_{h \rightarrow y}, W_{h \rightarrow 2}\right)$ are Gaussian.
		\item Homoskedastic (O) $Y$-noise, where $\sigma_y^2=e^2$ and $\sigma_2^2=1$, or heteroskedastic (E) $Y$-noise, where $\sigma_y^2=1$ and $\sigma_2^2=e^2$. 
	\end{enumerate} 
   To simplify the analysis, we consider four settings denoted by 1-4: POU, PEU, FOU, and FEU, where U represents the unscrambled $(X_1,X_2)$. Both $X_1$ and $X_2$ are generated as five-dimensional vectors and the environments parameter $e$ is set to $[0.2, 2.0, 5.0]$.
	
\begin{figure}[H]
	\centering  
	\begin{subfigure}{\textwidth}
		\center
		\includegraphics[width=0.245\textwidth, height=1 in]{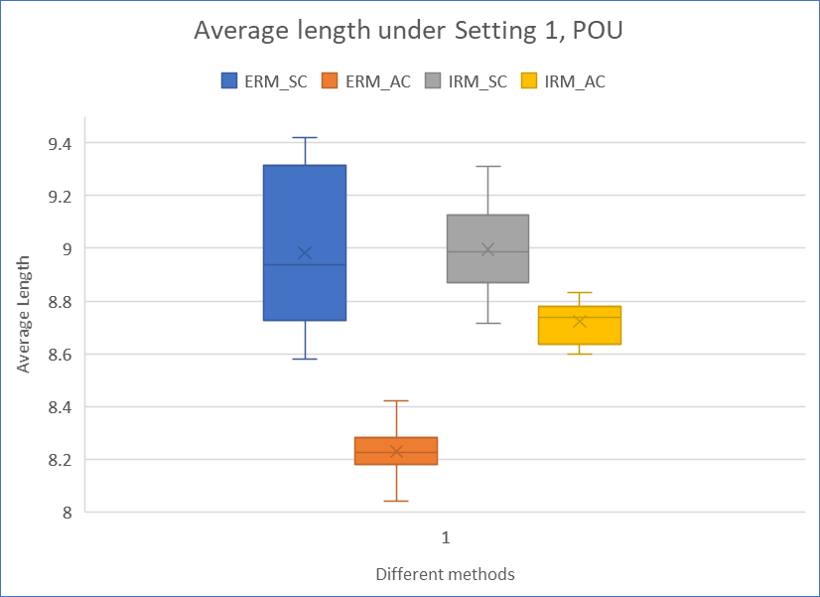}
		\includegraphics[width=0.245\textwidth, height=1 in]{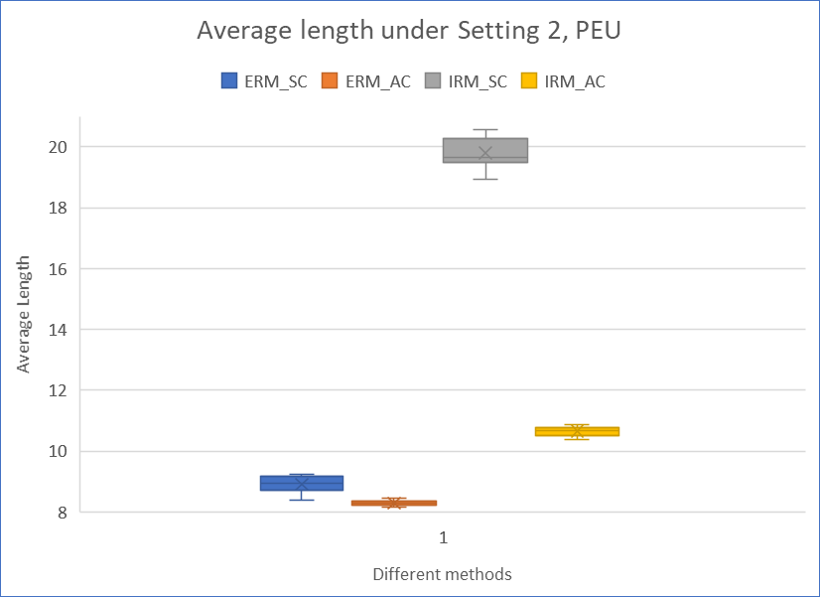}
		\includegraphics[width=0.245\textwidth, height=1 in]{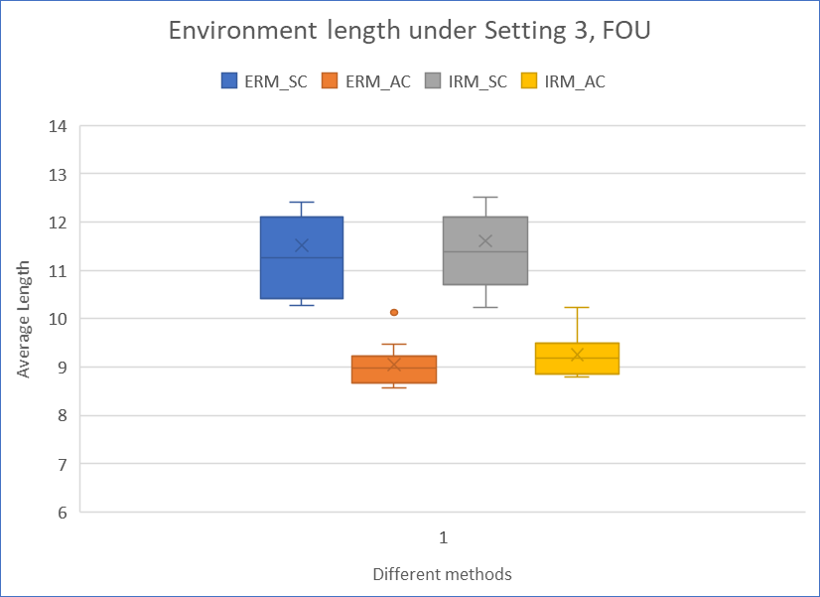}
		\includegraphics[width=0.245\textwidth, height=1 in]{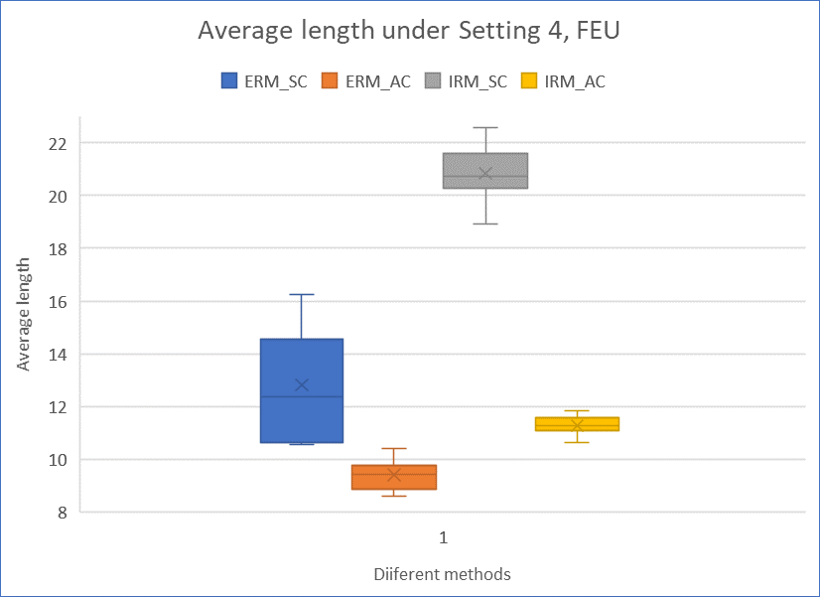}			
		\caption{Average length of different methods under different settings.}
		\label{Fig.sub.1}
	\end{subfigure}
	\vspace*{0.1cm}
	\begin{subfigure}{\textwidth}
		\center
		\includegraphics[width=0.245\textwidth, height=1 in]{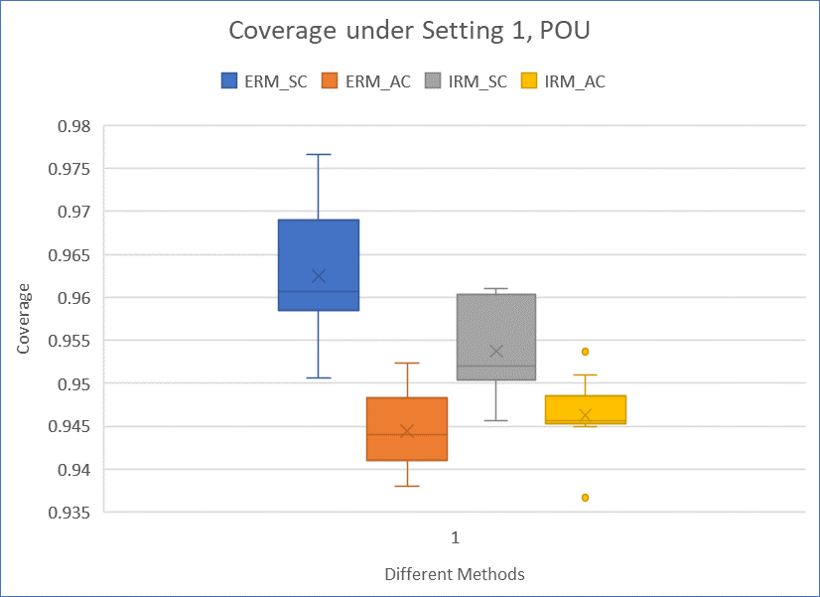}	
		\includegraphics[width=0.245\textwidth, height=1 in]{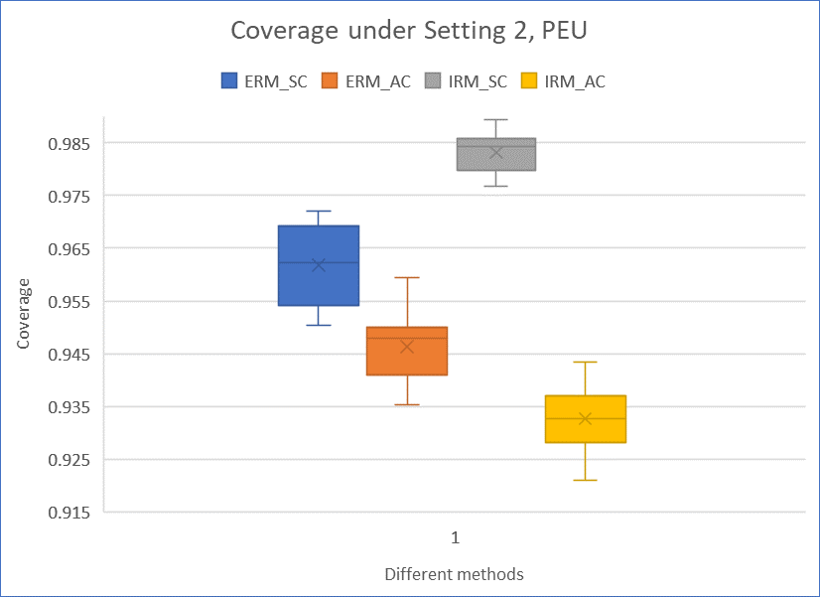}	
		\includegraphics[width=0.245\textwidth, height=1 in]{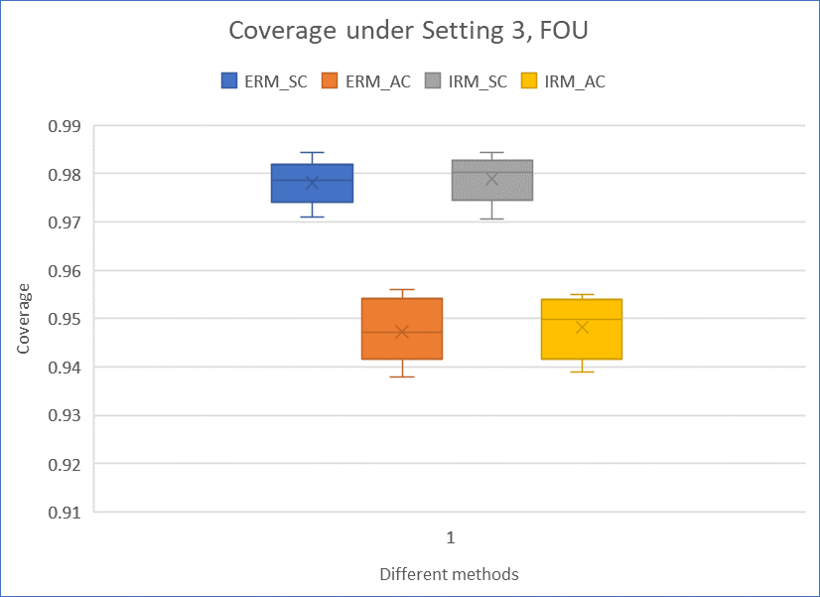}
		\includegraphics[width=0.245\textwidth, height=1 in]{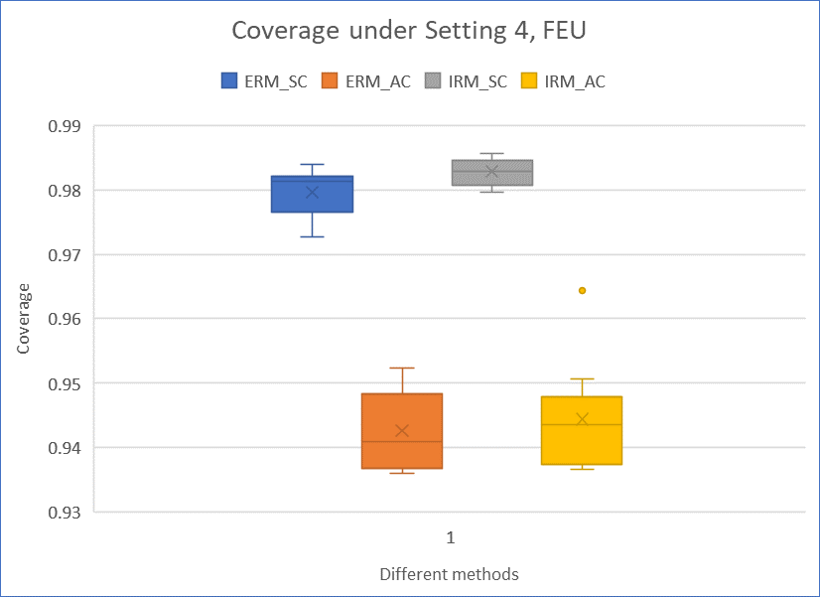}		
		\caption{Coverage of different methods under different settings.}
		\label{Fig.sub.2}
	\end{subfigure}
	\caption{The average length  and coverage value of SC and AC method on ERM and IRM models under different settings. Blue box is the SC interval for ERM estimates, Blue box is the SC interval for ERM estimates,,Blue box is the SC interval for ERM estimates, and Blue box is the SC interval for ERM estimates. }
	\label{Fig.main.1}
\end{figure}

\begin{figure}[H]
	\centering  
	\begin{subfigure}{\textwidth}
		\center
		\includegraphics[width=0.245\textwidth, height=1 in]{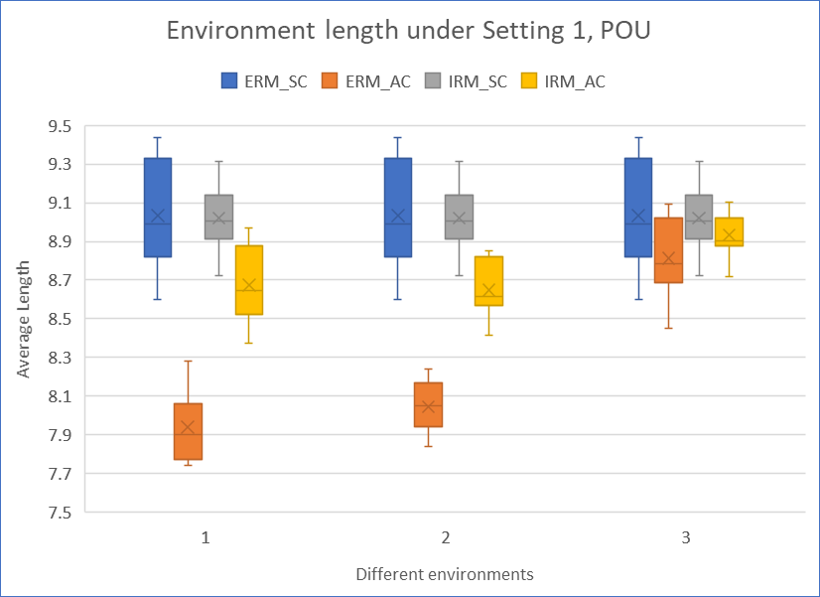}
		\includegraphics[width=0.245\textwidth, height=1 in]{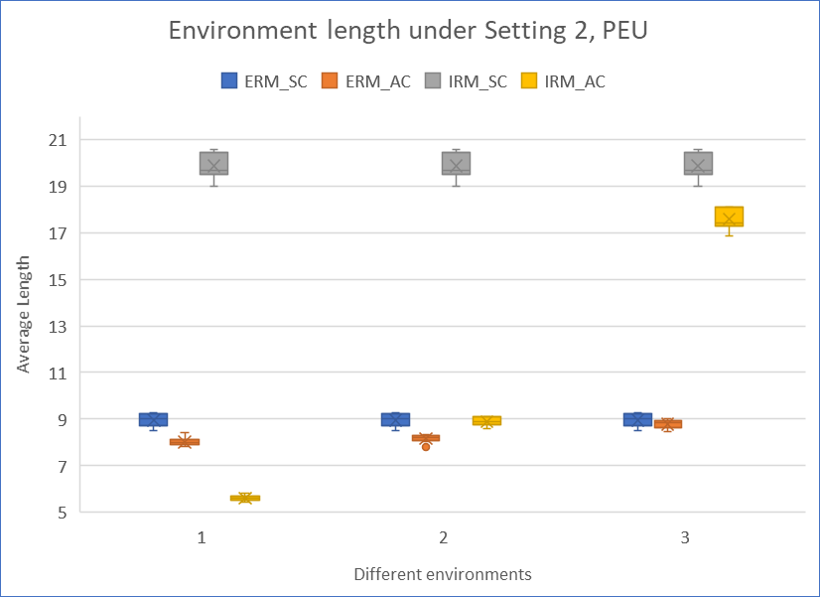}
		\includegraphics[width=0.245\textwidth, height=1 in]{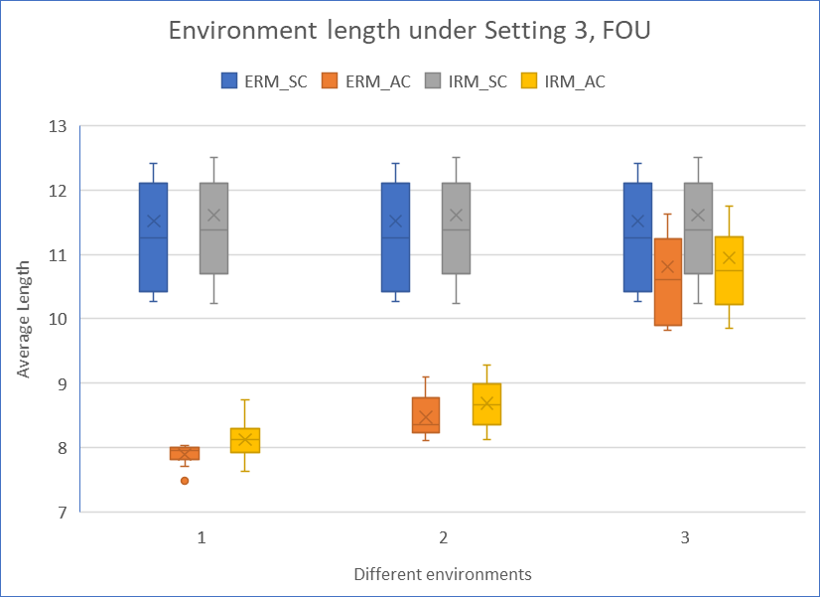}
		\includegraphics[width=0.245\textwidth, height=1 in]{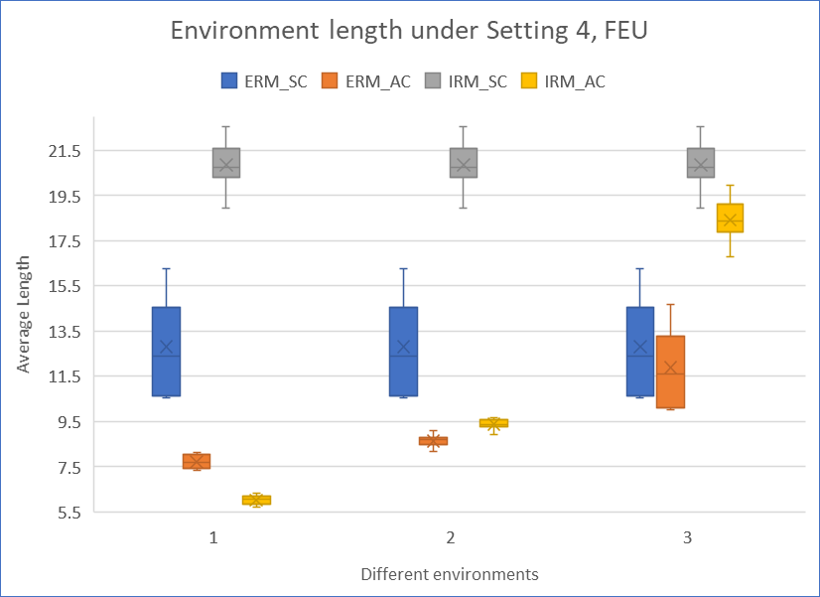}		
		\caption{Environment length of different methods under different settings.}
		\label{Fig.sub.21}
	\end{subfigure}
	\vspace*{0.1cm}
	\begin{subfigure}{\textwidth}
		\center
		\includegraphics[width=0.245\textwidth, height=1 in]{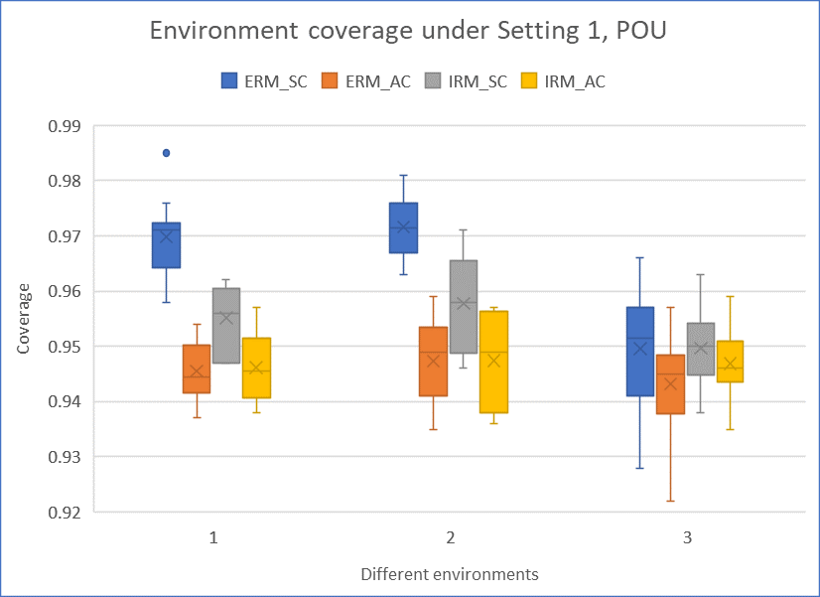}	
		\includegraphics[width=0.245\textwidth, height=1 in]{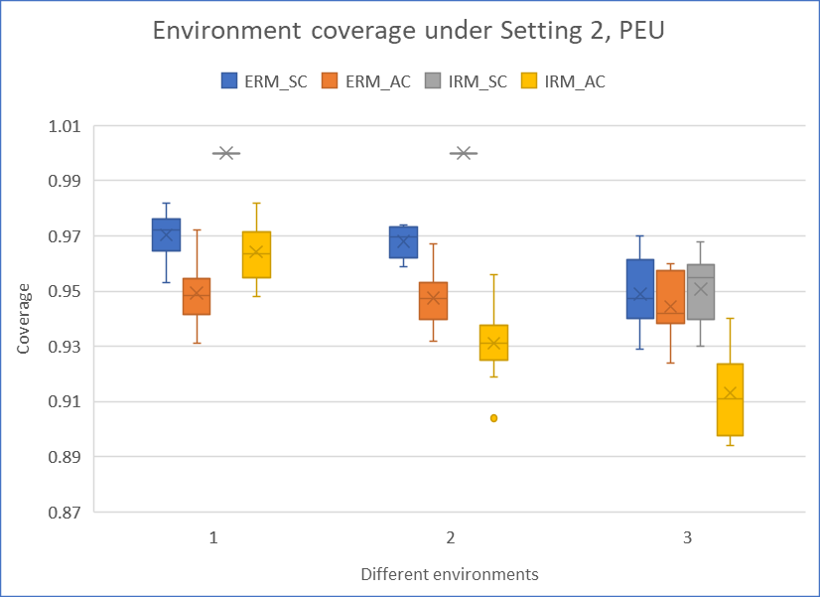}	
		\includegraphics[width=0.245\textwidth, height=1 in]{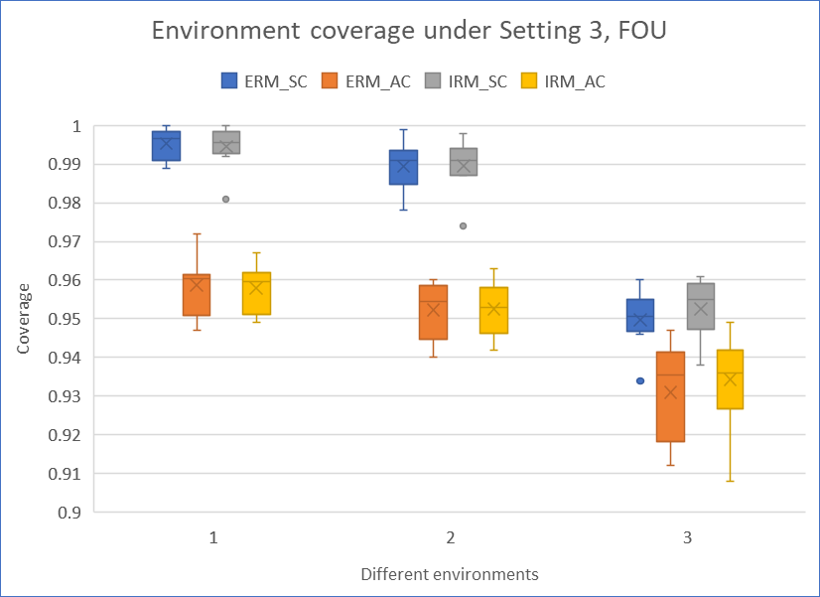}
		\includegraphics[width=0.245\textwidth, height=1 in]{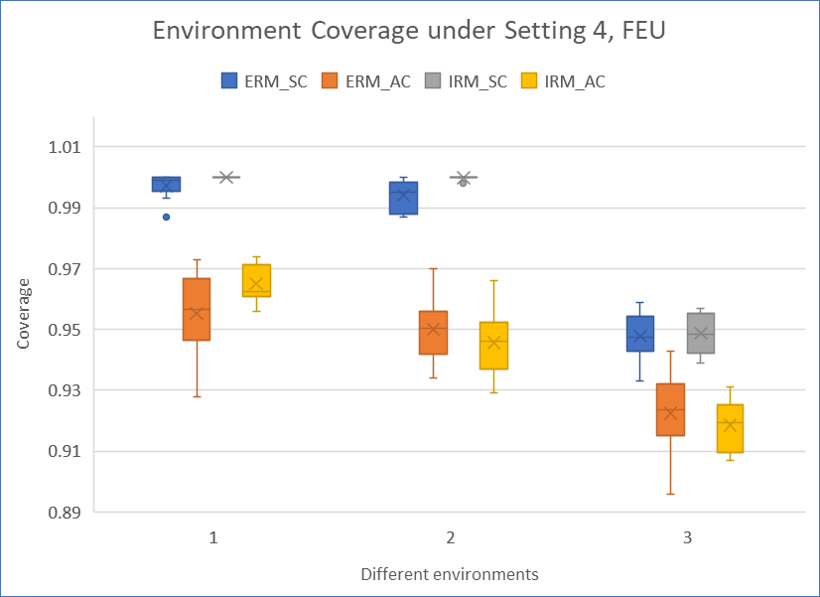}			
		\caption{Environment coverage of different methods under different settings.}
		\label{Fig.sub.22}
	\end{subfigure}
	\caption{The average length and coverage value of SC and AC method in each environment on ERM and IRM models under different settings. The labels are the same as Figure \ref{Fig.main.1}.  }
	\label{Fig.main.2}
\end{figure}

	Our method is applied to 6,000 independent observations from the described SEM model, with 2,000 used to train the IRMv1 and ERM models, and another 2,000 for calibration. The remaining data is reserved for testing. We construct split conformal intervals (SC) defined in (\ref{SC}) and adaptive conformal intervals (AC) defined in (\ref{ACIR}) at a miscoverage rate of $\alpha=0.05$. 
	The performance of the prediction intervals is evaluated on the test set using the following statistics: the average interval length is calculated by ${T}^{-1} \sum_{t=1}^T \tilde{Q}_{1-\alpha}(x_t)$, and the coverage rate calculated by ${T}^{-1} \sum_{t=1}^T \mathbbm{1}_{Y_t \in \hat{C}_{1-\alpha}\left(x_t\right)}$.
    In addition, we perform 20 replications by randomly splitting the training data 20 times and calculate the average statistics and their standard deviations to draw a boxplot. The performance of prediction intervals based on SC and AC under different settings is displayed in Figures \ref{Fig.main.1} and \ref{Fig.main.2}.

Figure \ref{Fig.main.1} shows the average performance on pooled data from all environments, while Figure \ref{Fig.main.2} displays the performance on each environment separately. 
In Figure \ref{Fig.sub.1}, the average length of the AC interval  is consistently smaller than that of the SC interval for both ERM and IRM estimates. Figure \ref{Fig.sub.2} illustrates that the average coverage of the AC interval remains stable under different settings, while the coverage in the SC method varies considerably among different settings. This suggests that, on average, the AC method achieves the desired coverage rate with smaller bandwidth under different settings compared to the SC method.

Figure \ref{Fig.sub.21} reveals that the length of the AC interval is adaptive to the environments, while the length of the SC interval is fixed, and the AC length is always smaller than the SC length in every environment. In Figure \ref{Fig.sub.22}, the coverage of the SC interval performs worse than the AC interval and results in an extreme coverage of $100\%$ in the heteroskedastic settings 2 and 4. While the coverage of the AC interval varies in different environments for the IRM estimates, it remains stable for ERM estimates. However, the coverage of the AC method performs better than the SC method for both IRM and ERM estimates, especially in the homoskedastic setting. This suggests that, in each environment, 
the AC method achieves the desired coverage rate with a narrower bandwidth under different settings compared to the SC method.


\subsection{Real data}
This section applies the ACIR method to financial data\footnote{https://www.kaggle.com/code/cnic92/explore-and-clean-financial-indicators-dataset/notebook} using the neural network setting in \cite{krueger2021out} and cleaned data consisting of factors in the U.S. stock market over five years,
resulting in 37 features of company basic information and a target variable representing the variation of stock prices.
The training data includes stock data from 2014 to 2016, while the testing data includes stock data from 2017 to 2018, with each year treated as an environment and varying sample sizes in each year. 

We begin by splitting the training data in each environment into equal calibration and proper training sets. We use the training set to fit the IRM model to obtain the invariant predictions. Then, we use the proposed method to construct adaptive conformal (AC) and split conformal (SC) prediction intervals for the stock price variation in 2017 and 2018 at a miscoverage rate of $\alpha=20\%$. The average length and coverage, as defined in Section 3.1, are computed and presented in Table \ref{real}. Additionally, we calculate the coverage and length in 2018 to compare the performance of the two methods in individual environments.
Table \ref{real} shows that both AC and SC achieve marginal coverage, while the AC method has a smaller length.

\begin{table}[H]
	\caption{The Performance of AC and SC method on financial data.}
	\label{real}
	\centering
	\begin{tabular}{ccccc}
		\toprule
		Method     & Coverage    & Length & Coverage in 2018 & Length in 2018\\
		\midrule
		SC &  $80.7\%$ & 21.5  &  $83.4\%$   &  21.5   \\
		AC  & $79.5\% $ & 17.4 &  $78.1\%$  &    15.4     \\
		\bottomrule
	\end{tabular}
\end{table}

	\section{Concluding remarks}
	This paper proposes an approach for uncertainty quantification of estimates obtained through Invariant Risk Minimization (IRM). To identify the environment to which a single test sample belongs, we propose a weight that indicates the similarity between the sample and the environment's sample space. This weight can be combined with environment-wise statistics, including environment-wise conformity scores, to obtain a weighted conformity score. We construct an adaptive conformal interval using the weighted conformity score, which is adaptive to distribution shifts across different environments. We demonstrate that the adaptive conformal interval enjoys a conditional coverage guarantee under conditions that can be checked by invariance assessment.

	The proposed method and its theoretical properties have a main limitation in that the weighted conformity score is defined by the first (mean) and second (standard deviation) moment similarity of the samples. This approach may not capture the characteristics of the sample distribution when using second moments, such as in the case of skewed data. To better capture the skewness, future research should consider alternative approaches that incorporate higher-order moments and are more reliable and reasonable.
	Additionally, while we propose a statistic for the assessment of invariance, it currently lacks theoretical and numerical details. Future work should consider using this assessment to calibrate the invariant representation and provide more thorough theoretical and numerical analysis of its performance.


	\bibliographystyle{apalike}
	\bibliography{ref.bib}


\end{document}